\documentclass[%
 aip,
 amsmath,amssymb,
 reprint,%
]{revtex4-1}

\usepackage{graphicx}% Include figure files
\usepackage{dcolumn}% Align table columns on decimal point
\usepackage{bm}% bold math
\usepackage[utf8]{inputenc}
\usepackage[T1]{fontenc}
\usepackage{mathptmx}
\usepackage{etoolbox}
\usepackage{caption}
\usepackage{subcaption}
\usepackage{verbatim}

\makeatletter
\def\@email#1#2{%
 \endgroup
 \patchcmd{\titleblock@produce}
  {\frontmatter@RRAPformat}
  {\frontmatter@RRAPformat{\produce@RRAP{*#1\href{mailto:#2}{#2}}}\frontmatter@RRAPformat}
  {}{}
}%
\makeatother
\begin{document}

\preprint{AIP/123-QED}

\title[Attractor reconstruction with reservoir computers]{Attractor reconstruction with reservoir computers: The effect of the reservoir's conditional Lyapunov exponents on faithful attractor reconstruction}
% Force line breaks with \\
\author{Joseph D. Hart}
 \email{joseph.hart@nrl.navy.mil}
\affiliation{ 
U.S. Naval Research Laboratory, Code 5675, Washington, DC 20375, USA%\\This line break forced with \textbackslash\textbackslash
}%

\date{\today}% It is always \today, today,
             %  but any date may be explicitly specified

\begin{abstract}

Reservoir computing is a machine learning framework that has been shown to be able to replicate the chaotic attractor, including the fractal dimension and the entire Lyapunov spectrum, of the dynamical system on which it is trained. We quantitatively relate the generalized synchronization dynamics of a driven reservoir during the training stage to the performance of the trained reservoir computer at the attractor reconstruction task. We show that, in order to obtain successful attractor reconstruction and Lyapunov spectrum estimation, the largest conditional Lyapunov exponent of the driven reservoir must be significantly more negative than the most negative Lyapunov exponent of the target system. We also find that the maximal conditional Lyapunov exponent of the reservoir depends strongly on the spectral radius of the reservoir adjacency matrix, and therefore, for attractor reconstruction and Lyapunov spectrum estimation, small spectral radius reservoir computers perform better in general. Our arguments are supported by numerical examples on well-known chaotic systems.
\end{abstract}

\maketitle

%\textbf{Reservoir computing is a machine learning framework that is known to be ... \cite{pathak2017using,lu2018attractor}.  }

\begin{quotation}
A common problem encountered in many areas of science occurs when a model of a dynamical system is required, but models based on first-principles are either unavailable, inaccurate, or intractable. In these cases, analysis must be developed based on time series data. Often, this is done by delay embedding techniques \cite{abarbanel2012analysis,kantz2004nonlinear}; however, machine learning approaches such as reservoir computing \cite{jaeger2002tutorial,jaeger2004harnessing} have shown significant promise in data-driven model development for dynamical systems \cite{brunton2022data}. 
Previous work \cite{pathak2017using,lu2018attractor} has shown that reservoir computing models can reproduce the dynamical system's Lyapunov spectrum, the most important quantifiers of chaotic dynamics. In this work, we address the issue of reconstructing the full Lyapunov spectrum of a dynamical system using reservoir computing. We identify the maximal conditional Lyapunov exponent of the untrained reservoir network during the training stage as a key metric for determining whether a reservoir computer can accurately reconstruct the Lyapunov spectrum of the target system.
\end{quotation}

%In this work, we show that the quality of a reservoir computer's performance of this Lyapunov spectrum estimation task is related to the untrained reservoir's dynamical response to the training signal through the theory of generalized synchronization and conditional Lyapunov exponents.}

%\section{Introduction}

Previous work has shown that a well-trained reservoir computer, when operated in an autonomous feedback mode, can replicate the attractor of the dynamical system on which it is trained, allowing it not only to produce accurate prediction of chaotic time series over five or more Lyapunov times \cite{pathak2017using}, but also to provide estimates of ergodic dynamical quantities such as the fractal dimension and entire Lyapunov spectrum of the target dynamical system \cite{pathak2017using,kobayashi2021dynamical,platt2021robust}. Perhaps most impressive is that reservoir computers were shown to be capable of providing accurate estimates of the negative Lyapunov exponents of the target dynamical system using only time series data, which is known to be a particularly difficult problem for other data-based Lyapunov exponent estimation techniques \cite{eckmann1985ergodic,sano1985measurement,zeng1991estimating,abarbanel2012analysis,kantz2004nonlinear}. Indeed, the reproduction of the Lyapunov spectrum has been suggested as a metric for evaluating a reservoir computer's performance at the attractor reconstruction task \cite{platt2021robust,platt2022systematic}. While other metrics of attractor reconstruction, such as valid prediction time \cite{jiang2019model,flynn2021multifunctionality}, are also commonly used, in this work we use the Lyapunov spectrum metric.

Despite this promise of reservoir computing for attractor reconstruction and estimating Lyapunov exponents from time series, Pathak et al. were unable to reproduce the negative Lyapunov exponent (and, therefore, the attractor dimension) of the Lorenz system, even though their reservoir computer accurately reproduced the positive and zero exponents and otherwise appeared to provide an accurate attractor reconstruction \cite{pathak2017using}. Pathak et al. attributed this failure to the especially thin transverse structure of the Lorenz return map, a peculiarity of the Lorenz system.

In this work, we provide evidence that the cause of this failure of the reservoir computer to reproduce the negative Lyapunov exponents is general and can be attributed %to the generalized synchronization dynamics of the reservoir network itself. We argue that, for successful Lyapunov spectrum replication, the reservoir contraction to the generalized synchronization manifold must be significantly faster than the contractions experienced by the true dynamical system. In other words, the maximal conditional Lyapunov exponent of the driven reservoir must be significantly more negative than the most negative Lyapunov exponent of the true system. The reason for this is closely related to the fact that an increase in fractal dimension can occur in filtered chaotic signals \cite{badii1988dimension}.
to the dynamics of the reservoir network itself. In the training phase, the reservoir network acts as a nonlinear filter driven by a chaotic signal, which is known to display a larger fractal dimension than the unfiltered chaotic signal when the filter's conditional Lyapunov exponents are less negative than the negative Lyapunov exponents of the drive system \cite{badii1988dimension,pecora1996discontinuous}. This increase in fractal dimension occurs because the filter cannot follow the rapid contractions corresponding to the most negative Lyapunov exponents of the drive system \cite{davies1996linear,davies1997reconstructing}. 

We argue that the magnitude of the maximal conditional Lyapunov exponent of the reservoir is predictive of whether the trained autonomous reservoir computer can replicate the negative Lyapunov exponents and the fractal dimension of the target chaotic system: For successful Lyapunov spectrum replication, the maximal conditional Lyapunov exponent of the driven reservoir must be significantly more negative than the most negative Lyapunov exponent of the target system. In other words, the reservoir contraction to a state of generalized synchronization with the drive system must be significantly faster than the contractions associated with the target system dynamics. %, otherwise the reservoir Lyapunov exponents will take the place of the true negative Lyapunov exponents in the Lyapunov spectrum. 
Our arguments are supported by numerical examples on well-known chaotic systems. 

We also find that the maximal conditional Lyapunov exponent of the reservoir depends strongly on the spectral radius of the reservoir adjacency matrix, suggesting that reservoirs with spectral radius signficantly smaller than one are better suited in general to the attractor reconstruction task than reservoirs with spectral radius near or greater than one. 

Our results not only provide insight into the key dynamical features that enable a reservoir computer to perform successful attractor reconstruction, but also provide an indicator for when one might expect that a trained reservoir computer has provided an accurate estimate of the negative Lyapunov exponents and attractor dimension of an unknown dynamical system.

%Provide an overview of the paper, section by section?

%different tasks need different reservoir conditions \cite{cramer2020control,jaegarForward}. In this work, we focus on the tasks of attractor reconstruction and Lyapunov exponent estimation. 

%add units to the LE figures.

\section{Reservoir computing}

A reservoir computer (RC) \cite{lukovsevivcius2009reservoir,lukovsevivcius2012reservoir} (here, we use an echo state network implementation \cite{jaeger2004harnessing}) is a type of recurrent neural network that is designed to be particularly easy to train. An RC consists of three layers: a linear input layer, a nonlinear recurrent layer (called the reservoir), and an output layer. Only the output layer is trained, allowing for computationally inexpensive training procedures such as ridge regression \cite{tikhonov1995numerical}. Reservoir computing has been shown to be well-suited to the performance of a variety of time-series analysis tasks, such as inferring unseen bifurcations, attractors, and tipping points \cite{gauthier2022learning,goldmann2022learn,kong2023reservoir,patel2023using}, causal inference \cite{banerjee2019using,banerjee2021machine}, chaos synchronization \cite{nazerian2023synchronizing} and communication \cite{antonik2018using}, and nonlinear control \cite{canaday2021model}  in a wide range of dynamical systems including delay systems \cite{goldmann2022learn}, spatio-temporal systems \cite{pathak2017using,pathak2018model,goldmann2022learn,kong2023reservoir} such as fluid turbulence \cite{nakai2018machine} and atmospheric dynamics \cite{arcomano2023hybrid,suematsu2022machine}, and real-world systems such as complex machinery \cite{thorne2022reservoir} and networks of neurons \cite{banerjee2023network}.

Each RC used here consists of a recurrent neural network of $N$ discrete-time nodes with random network topology (Erd{\"o}s-Renyi random networks with average degree 6). Let $\mathbf{r}[n]$ be an $N\times 1$ column vector that describes the state of the reservoir at time $n$. During training, the reservoir operates as a driven nonlinear dynamical system described by

\begin{equation}
\label{eq:reservoirEq}
\mathbf{r}[n+1] = 
\tanh{(A\mathbf{r}[n]+W^{in}\mathbf{x}[n]+\mathbf{b})},
\end{equation}
where $\mathbf{x}[n]$ is an $\mathcal{D}\times 1$ column vector describing the input signal to the reservoir, $A$ is an $N\times N$ matrix describing the inter-nodal connections in the reservoir layer, and $\mathbf{b}$ is an $N\times 1$ bias vector with elements chosen randomly and uniformly between -1 and 1. An important quantity will be the spectral radius of $A$, denoted by $\rho$, which can be changed by scaling all elements of $A$. As in Refs. \cite{pathak2017using,lu2018attractor}, the $N\times\mathcal{D}$ matrix $\tilde{W}^{in}$ is chosen randomly so that each row has
one non-zero element, chosen uniformly between –1 and 1. The input matrix $W^{in}$ is then determined by $W^{in}=\sigma_{in}\tilde{W}^{in}$. In this work, we consider input signals that are created by sampling a continuous-time dynamical system described by the state vector $\mathbf{x}(t)$, such that $\mathbf{x}[n]\equiv\mathbf{x}(n\tau)$ where $\tau$ is the uniform sampling time.

For the task of attractor reconstruction, the reservoir is driven by the input signal $\mathbf{x}[n]$ in the training phase. The reservoir state $\mathbf{r}[n]$ is recorded at each time step after some initial transient time, and the RC is trained to predict that signal at the next time step $\mathbf{x}[n+1]$; this prediction $\hat{\mathbf{x}}$ is obtained from the reservoir by

\begin{equation}
\label{eq:xhat}
\hat{\mathbf{x}}[n] = W^{out}\mathbf{P}(\mathbf{r}[n]),    
\end{equation}
where $\mathbf{P}$ is an $N_p$-dimensional function of $\mathbf{r}$, and $W^{out}$ is a $\mathcal{D}\times N_p$ matrix obtained by ridge regression \cite{tikhonov1995numerical}. Often, the choice $P(\mathbf{x})=\mathbf{x}$ is made. Following Refs. \cite{pathak2017using} and \cite{platt2022systematic}, we do not normalize the input variables to the reservoir computer or reservoir variables when performing the ridge regression.

Once the training is complete, attractor reconstruction can be attempted by feeding the prediction $\hat{\mathbf{x}}$ back into the reservoir as the input, turning the reservoir into an autonomous dynamical system:
\begin{equation} \begin{split}
\label{eq:autonomous_RC}
\mathbf{r}[n+1] &= \tanh{(A\mathbf{r}[n]+W^{in}\hat{\mathbf{x}}[n]+\mathbf{b})}.
\end{split} \end{equation}

Attractor reconstruction does not always succeed, even when the one-step-ahead training error is very small. Often, when attractor reconstruction fails, the RC approaches an untrained attractor, such as a fixed point or a limit cycle \cite{lu2018attractor}.  

When attractor reconstruction does succeed, it has been shown that the autonomous RC described by Eq. \ref{eq:autonomous_RC} can provide a stable reconstruction of the attractor of the target dynamical system as quantified by density distributions, Poincar{\'e} sections, Lyapunov exponents, and information dimension \cite{pathak2017using,lu2018attractor,kobayashi2021dynamical}. However, it is known that in some cases, even though the density distributions and Poincar{\'e} sections of the RC output and the target system are essentially indistinguishable, the negative Lyapunov exponents (and therefore the Kaplan-Yorke dimension) of the target system are not reproduced by the autonomous reservoir, while in other cases a large number of the target system negative Lyapunov exponents are reproduced with remarkable accuracy \cite{pathak2017using}. In this work, we shed light on when RCs fail to reproduce the negative Lyapunov exponents of the target system, and we discuss how one can tune a RC to be more likely to reproduce the negative Lyapunov exponents and therefore the Kaplan-Yorke dimension of the target system.

\section{Conditional Lyapunov exponents and reservoir computing}

In this section, we review the concept of the conditional Lyapunov exponents (CLEs) of driven systems \cite{pecora1991driving} and their relationship to generalized synchronization \cite{abarbanel1996generalized,kocarev1996generalized}. We then review the well-known result that an increase in attractor dimension occurs in filtered dynamical systems when the maximal CLE of the filter dominates the negative Lyapunov exponents of the driving system \cite{badii1988dimension} because in this situation the filter cannot follow the rapid contractions of the driving system \cite{davies1996linear,davies1997reconstructing}. We apply these concepts to reservoir computing by noting that, in the training stage, the reservoir is a nonlinear filter, and therefore its maximal CLE plays a key role in determining the attractor dimension of the reservoir.  Finally, we argue that the CLEs of the reservoir in the training stage affect the performance of the reservoir computer at the attractor reconstruction task.

\subsection{Conditional Lyapunov exponents \label{sec:CLEs}}

Conditional Lyapunov exponents arise in the context of driven dynamical systems \cite{pecora1991driving}. By a ``driven dynamical system,'' we mean that two systems (the ``drive system'' and the ``response system'') are unidirectionally coupled, such that the drive system affects the response system, but the response system does not affect the drive system. The full system (consisting of the drive system $\mathbf{d}[n]$ and response system $\mathbf{z}[n]$) can be described by a set of $(\mathcal{D}+\mathcal{D}_r)$ equations, where $\mathcal{D}$ is the dimension of the drive system and $\mathcal{D}_r$ is the dimension of the response system:

\begin{equation}
    \label{eq:drive_response}
    \begin{split}
        \mathbf{d}[n+1] &= \mathbf{F}(\mathbf{d}[n]) \\
        \mathbf{z}[n+1] &= \mathbf{G}(\mathbf{d}[n],\mathbf{z}[n]).
    \end{split}
\end{equation}

In order to quantify the stability of the response system's dynamics to the drive, consider an infinitesimal perturbation to the response system $\delta \mathbf{z}$. This leads to the following variational equation:

\begin{equation}
    \label{eq:genVarEq}
    \delta \mathbf{z}[n+1] = D\mathbf{G}(\mathbf{d}[n],\mathbf{z}[n])\delta\mathbf{z}[n],
\end{equation}
where $D\mathbf{G}$ is the Jacobian function of $\mathbf{G}$ with respect to $\mathbf{z}$. If all $\mathcal{D}_r$ Lyapunov exponents of Eq. \ref{eq:genVarEq} are negative, then the trajectory of the response system is stable to small perturbations. These Lyapunov exponents depend on the drive \textbf{d}[n], and are therefore called conditional Lyapunov exponents (CLEs) \cite{pecora1991driving}. We denote the CLEs of the response system as $\lambda_i^{(r)}$.

The dynamics of the reservoir in the training stage are described by Eq. \ref{eq:reservoirEq}. We can make the associations that $\mathbf{d}[n]=\mathbf{x}[n]$ and $\mathbf{z}[n]=\mathbf{r}[n]$. The CLEs of the reservoir to the drive signal $\lambda^{(r)}_i$ are computed as the Lyapunov exponents of the variational equation associated with Eq. \ref{eq:reservoirEq} that considers perturbations to $\mathbf{r}$ only:

\begin{equation}
    \label{eq:conditionalVarEq}
    \delta r_i[n+1] = \mathrm{sech}^2{\big(A_{ik}r_k[n] + W^{in}_{i\ell}x_\ell[n]}+b_i\big)A_{ij}\delta r_j[n],
\end{equation}
where summation over repeated indices is implied. In cases where the training data is limited and in which attractor reconstruction is successful, one can replace $x_\ell[n]$ in Eq. \ref{eq:conditionalVarEq} with $\hat{x}_\ell[n]$ to obtain an estimate of the CLEs. A similar technique was used for Lyapunov exponent estimation in Refs. \cite{lu2018attractor,hart2023estimating}.

\subsection{Generalized synchronization}

Generalized synchronization between two dynamical systems (often a drive system and a response system) occurs when there is a functional relationship between the dynamical variables of the two systems \cite{rulkov1995generalized,kocarev1996generalized,abarbanel1996generalized}. A standard test for generalized synchronization is the auxiliary system approach \cite{abarbanel1996generalized}: Two identical copies of the response system with different initial conditions synchronize with each other when driven by the same drive signal if and only if they display generalized synchronization with the drive signal. The CLEs quantify the rate at which the responses of the two identical systems converge to the same trajectory \cite{pecora1991driving}. Generalized synchronization is stable if and only if all CLEs are negative \cite{abarbanel1996generalized,kocarev1996generalized}.

Generalized synchronization and the closely related concept of consistency \cite{uchida2004consistency} of the reservoir response to the drive signal have been identified as essential for a RC to be effective and reliable \cite{lu2018attractor,lymburn2019reservoir,hart2020embedding,pyragas2020using,grigoryeva2021chaos,larger2012photonic,nakayama2016laser,lymburn2019consistency,platt2021robust,platt2022systematic}. In particular, it is known that if the generalized synchronization function between the reservoir and the drive signal is one-to-one and smooth, then the Lyapunov spectrum of an ideal autonomous reservoir computer can reproduce the full Lyapunov spectrum of the target system \cite{lu2018attractor}. Despite this extensive literature, only a few works have considered the impact of the magnitude of the reservoir's maximal CLE on any type of reservoir computing task \cite{lymburn2019consistency,lymburn2019reservoir,platt2021robust}. As far as we are aware, none have linked the magnitude of the maximal CLE of the reservoir to the reservoir's performance at the attractor reconstruction task, as we do in the following sections.

%The full system has $(\mathcal{D}+\mathcal{D}_r)$ Lyapunov exponents; $\mathcal{D}$ of these are given by the Lyapunov exponents of the drive system, which are unchanged by the presence of the response, and are therefore denoted by $\lambda_i^{(drive)}$. The other $\mathcal{D}_{r}$ Lyapunov exponents of the full system depend on the drive \textbf{d}[n], and are therefore denoted by $\lambda_i^{(filter)}$ and called ``conditional Lyapunov exponents'' \cite{pecora1991driving}. 

\subsection{Attractor dimension increase in filtered dynamical systems \label{sec:filtering}}
It is well-established that an increase in dimension can occur in dynamical signals that are filtered by an infinite impulse response (IIR) filter (also called a recursive filter) \cite{badii1988dimension}. In this subsection, we briefly review this result, and relate it to the CLEs of the filter.

For simplicity of presentation we consider an ideal linear low-pass filter that is driven by one variable of a chaotic system (as done by Badii et al \cite{badii1988dimension}), but the results in this section also apply to higher-order filters \cite{davies1996linear,davies1997reconstructing}. We consider a driven, lowpass filter:

\begin{equation}
\label{eq:linearFilt}
    \dot{z}_{f}(t) = -\eta z_{f}(t) + u(t). 
\end{equation}
In Eq. \ref{eq:linearFilt}, $z_{f}$ is the filter output and $u(t)$ is the filter input, which in this simple example is one variable of a chaotic system with $\mathcal{D}$ variables (e.g., the $x$ variable of the Lorenz system). The filter cut-off frequency $\eta$ is taken to be positive so that the filter is stable. This filtered dynamical system can be written in the form of Eq. \ref{eq:drive_response} and can therefore be considered a drive-response system. The $\mathcal{D}$ Lyapunov exponents associated with the drive $\lambda_i^{(drive)}$ are unaffected by the presence of the filter. In this special case of a linear filter, the Lyapunov exponent associated with the filter response to the drive $\lambda^{(filter)}=-\eta$ is independent of the drive. In the formalism presented in Section \ref{sec:CLEs}, $\lambda^{(filter)}$ is the conditional Lyapunov exponent of the filter. In a stable linear filter, the filter always displays generalized synchronization with the drive. %A reconstruction of the dynamics of the full system (drive and filter response) can be obtained by a time-delay embedding of the filter output $z_{f}(t)$ \cite{badii1988dimension}.

The fractal dimension of a chaotic system is given by the Kaplan-Yorke conjecture as $\mathcal{D}_{KL}=j+\sum_{k=1}^j\lambda_j/|\lambda_{j+1}|$, where $j$ is the largest index for which the sum $\sum_{k=1}^j\lambda_k$ is non-negative \cite{kaplan1979functional}. If $\lambda^{(filter)}\leq\lambda_j^{(drive)}$, then $\lambda^{(filter)}$ does not factor into the dimension calculation, and $\mathcal{D}_{KL}$ of the filtered chaotic system is the same as $\mathcal{D}_{KL}$ of the unfiltered chaotic system. However, if $\lambda^{(filter)}>\lambda_j^{(drive)}$, then $\lambda^{(filter)}$ replaces $\lambda_j^{(drive)}$ in the computation of $\mathcal{D}_{KL}$, causing $\mathcal{D}_{KL}$ of the filtered chaotic system to be greater than $\mathcal{D}_{KL}$ of the unfiltered chaotic system. In some cases, $\mathcal{D}_{KL}$ of the filtered chaotic system can be significantly greater than of the unfiltered system \cite{badii1988dimension,pecora1996discontinuous,carroll2020dimension}.% In Badii's original paper, it was pointed out that this dimension increase could be problematic in situations when one is attempting to reconstruct the dynamics of the driving chaotic system using a time-delay embedding of the filter output $z_{f}(t)$, as might be the case if one were using a lowpass filter to remove noise from an experimental measurement \cite{badii1988dimension}.

\subsubsection{Time scales}

Another way of thinking about increase in attractor dimension due to filtering is in terms of time scales \cite{davies1997reconstructing}. If $\lambda^{(filter)}<\lambda_j^{(drive)}$, ($j$ is still the largest index for which the sum $\sum_{k=1}^j\lambda^{(drive)}_k$ is non-negative) the average phase space contractions of the drive system are slow relative to the filter response, and the filter variable is able to closely follow the drive dynamics. The filter is approximately isochronally synchronized to the drive system, and therefore the presence of the filter variable does not affect the dimensionality of the combined system. On the other hand, if $\lambda^{(filter)}>\lambda_j^{(drive)}$, the average phase space contractions of the drive system are faster than the filter is able to respond, and the filter dynamics play a nontrivial role in the combined system dynamics. Therefore, the filter variable does contribute to the combined system dimensionality. 

These concepts were formalized for linear filters by Davies \cite{davies1997reconstructing} in the language of generalized synchronization and embedding theorems. Since the filter is assumed stable, $\lambda^{(filter)}<0$ and the filter displays generalized synchronization with the drive. This means that there is a function that maps the drive phase space to the filter phase space.

If $\lambda^{(filter)}<\lambda_j^{(drive)}$, the function that maps the drive phase space to the filter phase space is continuous and smooth \cite{davies1997reconstructing}. On the other hand, if $\lambda^{(filter)}>\lambda_j^{(drive)}$, the filter and drive still display generalized synchronization, but the functional relationship is no longer smooth, and the dynamics of the drive system cannot be fully reconstructed by Takens embedding from the dynamics of the filter \cite{davies1997reconstructing}. 

\subsection{Application to reservoir computing}\label{sec:apptoRC}

%\subsection{The reservoir as a nonlinear filter}
While the arguments presented above for dimension increase due to IIR filtering were for a simple linear filter, the results hold for more complex linear IIR filters and for nonlinear recursive filters \cite{badii1988dimension,davies1996linear,davies1997reconstructing}. If the filter is linear, then the filter Lyapunov exponent(s) are independent of the drive signal. However, if the filter is nonlinear, the filter Lyapunov exponent(s) depend on the drive signal and are therefore CLEs \cite{pecora1991driving}. It is the relation between the filter CLEs and the drive Lyapunov exponents that determine whether there is an attractor dimension increase in filtered dynamical systems.

In the training stage, the driven (non-autonomous) reservoir can be thought of as a nonlinear recurrent filter. %Unlike the case of linear filters, for nonlinear filters the filter Lyapunov exponents depend on the drive signal itself, and are therefore CLEs \cite{pecora1991driving}. 
In the training stage (Eq. \ref{eq:reservoirEq}), the complete reservoir-and-drive system (considered as a whole) has $\mathcal{D}+N$ Lyapunov exponents, where $\mathcal{D}$ is the dimension of the drive system. These $\mathcal{D}+N$ Lyapunov exponents consist of the $\mathcal{D}$ exponents of the drive system (which are unaffected by the presence of the reservoir) and the $N$ CLEs of the reservoir, which we label $\lambda^{(drive)}_i$ and $\lambda^{(r)}_i$, respectively.

%**********This is the reference to use to prove this. Maybe write it up this weekend in the supplement and mention it in the general section regarding CLEs\cite{davies1996linear, davies1997reconstructing}.

%***********Say that we hypothesize that Davies' results hold for nonlinear filters, with the substitution that the nonlinear filter CLEs are an analog of the linear filter Lyapunov exponents. (IF THIS IS NOT ESSENTIAL MAYBE LEAVE IT OUT AND SAVE IT FOR LATER).

If the maximal CLE of the reservoir $\lambda_{max}^{(r)}$ is negative, then the reservoir displays generalized synchronization to the drive system. According to the arguments in Section \ref{sec:filtering}, if $\lambda_{max}^{(r)}$ is less than the most negative Lyapunov exponents of the drive system, then the reservoir is able to respond quickly enough to follow the dynamics (including the rapid contraction dynamics associated with the negative Lyapunov exponents) of the drive system. On the other hand, if $\lambda_{max}^{(r)}>\lambda^{(drive)}_i$, then the reservoir cannot respond quickly enough to follow the contractions associated with $\lambda^{(drive)}_i$. 

In the literature, this slower reservoir response is often interpreted positively, as it can be considered to endow the reservoir with memory \cite{goldmann2020deep}. However, in the context of attractor reconstruction, it means that $\lambda_{max}^{(r)}$ takes the place of the target system Lyapunov exponent in the Kaplan-Yorke computation, and therefore the Kaplan-Yorke dimension of the reservoir dynamics is greater than the Kaplan-Yorke dimension of the drive system. In the terminology of generalized synchronization, the generalized synchronization function between the reservoir and the drive signal still exists but it is not smooth, and, therefore, the dynamics of the drive system can not be fully reconstructed by Takens embedding from the reservoir dynamics \cite{davies1997reconstructing}. 

%*******Additionally, if our hypothesis that Davies' results can be extended to nonlinear filters is correct, then when $\lambda_{max}^{(c)}>\lambda^{(drive)}_i$, the functional relationship between the drive system phase space and the reservoir phase space is not smooth, and therefore the full original state space cannot be reconstructed from a Takens embedding of the reservoir state space. (IF THIS IS NOT ESSENTIAL, MAYBE LEAVE IT OUT AND SAVE IT FOR LATER).

\subsection{Application to attractor reconstruction}

For the tasks of attractor reconstruction and Lyapunov exponent estimation, the RC is operated in autonomous mode (Eq. \ref{eq:autonomous_RC}), in which the RC output is fed back and used as the input to the reservoir. Since the system is autonomous (and not driven), there is no concept of CLEs. There are $N$ total Lyapunov exponents, which we label $\lambda_i^{(RC)}$.

When attractor reconstruction is successful, the RC output is a good approximation of the true driving system \cite{pathak2017using,lu2018attractor}. In this case, one might expect some of the $\lambda_i^{(RC)}$ to replicate the target Lyapunov exponents while the other $\lambda_i^{(RC)}$ stay close to the CLEs of the driven, non-autonomous reservoir. We find that this is often the case. We therefore propose to use $\lambda^{(r)}_{max}$ as an estimate of the maximal Lyapunov exponent of the autonomous RC that should be attributed to the reservoir dynamics rather than to the target dynamical system.

For successful attractor reconstruction, the RC should replicate the ergodic properties of the dynamical system on which it is trained, including the Lyapunov exponents and the information dimension \cite{platt2021robust,platt2022systematic}. According to the arguments in Sections \ref{sec:filtering} and \ref{sec:apptoRC}, this is possible only if all of the reservoir CLEs $\lambda_i^{(r)}$ are more negative than the most negative Lyapunov exponent of the target dynamical system. As shown below, we find the best performance when the reservoir CLEs are significantly more negative than the target negative exponents.

We note that even when attractor reconstruction does succeed and the reconstructed attractor is locally stable, the autonomous RC can exhibit multistability in which untrained attractors can be reached from certain initial conditions \cite{flynn2021multifunctionality,flynn2023seeing}. In this work, we assume that the RC dynamics are on the trained attractor. Since the CLE and LE spectra are properties of a specific attractor, we do not expect the potential presence of multiple attractors to affect these quantities.

Much attention is paid to the spectral radius of the reservoir. Conventional wisdom is that the spectral radius of the RC should be near 1 \cite{jaeger2002tutorial,verstraeten2009quantification}, as this is thought to provide a good trade off between long memory and stability (generalized synchronization). However, ``memory'' in terms of reservoir computing is often considered to be the inverse of the maximal CLE $\lambda_{max}^{(r)}$: a larger maximal CLE leads to greater memory \cite{carroll2020dimension}. According to the arguments above, this ``memory'' can apparently be detrimental for the task of attractor reconstruction (defined as successful when the Lyapunov exponents and dimension of the trained autonomous RC agrees with the true Lyapunov exponents and attractor dimension), and the reservoir spectral radius should often be significantly smaller than 1 for this task, as was noted in Ref. \cite{hart2023estimating}. Other definitions of successful attractor reconstruction, such as accurate short- or medium-term prediction error, can result in different optimal values for the reservoir spectral radius \cite{jiang2019model,flynn2021multifunctionality,flynn2023seeing,jaurigue2024chaotic}.

In the following sections, we provide evidence for these arguments by performing attractor reconstruction on the Lorenz and Qi chaotic systems using RCs with a variety of different spectral radii. We find that a well-trained autonomous RC replicates the Lyapunov spectra and Kaplan-Yorke dimension of the target dynamical system only when $\lambda_{max}^{(r)}$ is sufficiently small and that $\lambda_{max}^{(r)}$ is strongly correlated with the spectral radius of the reservoir adjacency matrix.

\section{Results: Lorenz system}
\subsection{The Lorenz Equations}

The Lorenz system is described by \cite{lorenz1963deterministic}
\begin{equation}
\label{eq:lorenz}
\begin{array}{l}
\frac{{dx}}{{dt}} =\sigma\left( {y - x} \right) + a\xi_x(t)\\
\frac{{dy}}{{dt}} = x \left( \rho - z\right) - y + a\xi_y(t)\\
\frac{{dy}}{{dt}} = xy - \beta z + a\xi_z(t).
\end{array}
\end{equation}
In this work, we use $\sigma=10$, $\beta=8/3$ and $\rho=28$ or $\sigma=20$, $\beta=4$, and $\rho=45.92$. The first set of parameters is the standard set; the second has been shown to have a negative Lyapunov exponent with large magnitude \cite{story2009application}. The dynamical noise $\xi(t)$ is Gaussian white noise with unit standard deviation and the different $\xi_i$ are independent. We choose the dynamical noise strength $a=0.01$; dynamical noise is known to help with the RC training in some cases \cite{banerjee2019using, rohm2021model,banerjee2021machine}. The equations were numerically integrated using a second order Runge-Kutta method with the noise handled in the method of Honeycutt \cite{honeycutt1992stochastic} with an integration time step of 0.001. The time series obtained by the integration were downsampled to have $\tau$ = 0.02, as in Refs.  \cite{pathak2017using,hart2023estimating}. All three variables of the Lorenz system are used as inputs to the reservoir.

We consider two different values for the Lorenz parameters. We compute the Lorenz Lyapunov spectra to be 0.91, 0.0, -14.6 ($\sigma=10$) and  1.5, 0.0, and -26.5 ($\sigma=20$) in agreement with Refs. \cite{pathak2017using,story2009application}.

\subsection{Attractor reconstruction}

\begin{figure}
    \centering
    \begin{subfigure}[b]{0.45\textwidth}
    \includegraphics[width=\textwidth]{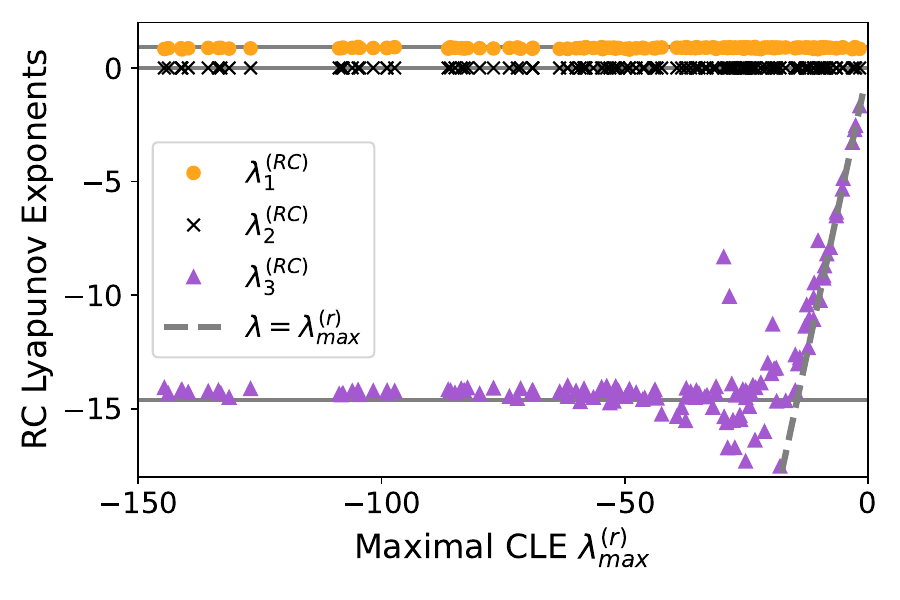}
    \caption{}
    \end{subfigure}
    \begin{subfigure}[b]{0.45\textwidth}
    \includegraphics[width=\textwidth]{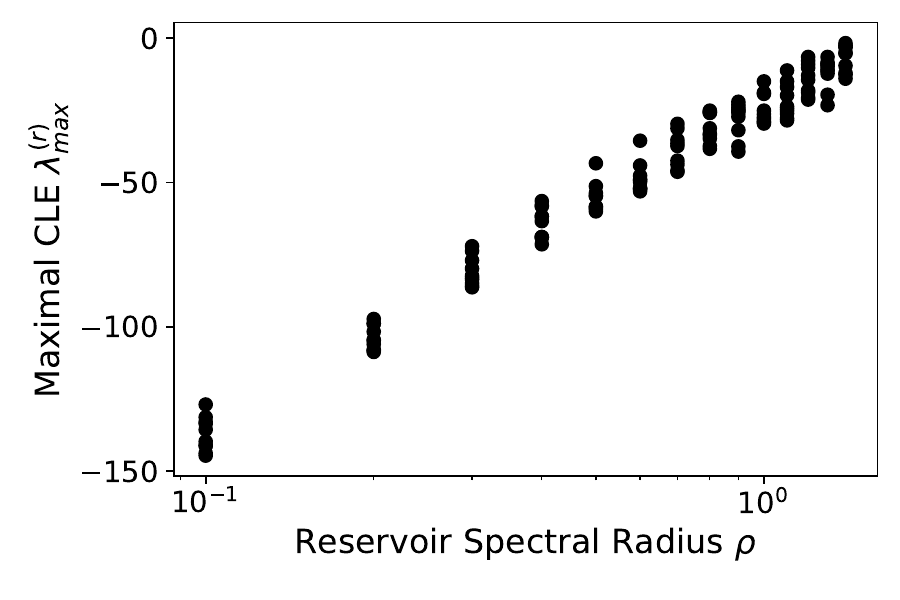}
    \caption{}
    \end{subfigure}
    \begin{subfigure}[b]{0.45\textwidth}
    \includegraphics[width=\textwidth]{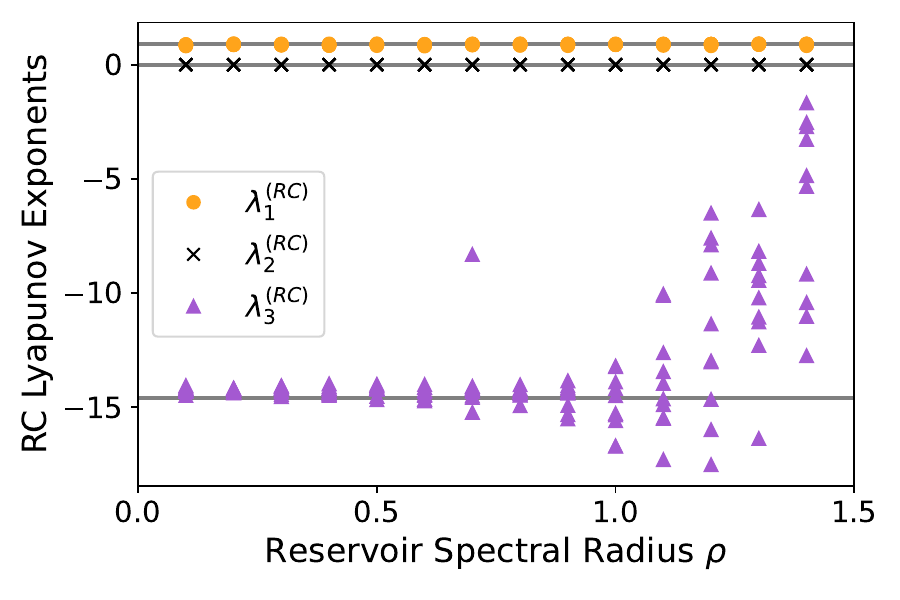}
    \caption{}
    \end{subfigure}
    \caption{Lorenz system with standard parameters: $\sigma=10$, $\beta=8/3$ and $\rho=28$. (a) Three largest Lyapunov exponents of the trained autonomous RC $\lambda_i^{(RC)}$ vs the maximal CLE of the driven reservoir $\lambda_{max}^{(r)}$. The Lyapunov exponents of the target Lorenz system are indicated as solid gray lines. The line $\lambda=\lambda_{max}^{(r)}$ is shown as a dashed gray line. (b) Maximal CLE of the driven reservoir $\lambda_{max}^{(r)}$ vs the reservoir spectral radius. (c) Three largest Lyapunov exponents of the trained autonomous RC $\lambda^{(RC)}$ vs the reservoir spectral radius. }
\end{figure}

For each value of the reservoir spectral radius, we create, train, and test 1000 RCs with $N=300$ nodes. Following Ref. \cite{pathak2017using}, the reservoir output $\hat{\mathbf{x}}=[\hat{x},\hat{y},\hat{z}]^T$ is given by

\begin{equation}
    \begin{bmatrix}
        \hat{x}[n] \\ \hat{y}[n] \\ \hat{z}[n]
    \end{bmatrix} = W^{out}\begin{bmatrix}\mathbf{r}[n] \\
        \mathbf{r}[n] \\
        \tilde{\mathbf{r}}[n] 
    \end{bmatrix},
\end{equation}
where $\tilde{\mathbf{r}}$ is defined such that  the first half of its elements are the same as that of $\mathbf{r}$, while $\tilde{\mathbf{r}}= r^2$ for the remaining half of the reservoir nodes.  The RCs differ in their input strengths $\sigma_{in}$ (chosen randomly in the range (0,0.2]), random network topologies (Erd{\"o}s-Renyi random networks with average degree 6), input matrices, and bias vectors. We use 40000 training time steps and set the ridge regression parameter to zero for the training (so that only simple regression is used); the noise is sufficient to prevent overfitting \cite{bishop1995training,wikner2023stabilizing}. The ten ``best'' attractor reconstructions are retained, where the attractor reconstructions are ranked according to the mean square error between the power spectra of the reconstructed $x$ variable and of the true $x$ variable, computed over 20000 time steps. The Lyapunov spectra of the trained, autonomous RCs $\lambda_{i}^{(RC)}$ are computed using 100000 time steps, and the CLE spectra of the driven reservoirs $\lambda_{i}^{(r)}$ are computed over the 40000 training time steps.

First, we consider the attractor reconstruction task on the Lorenz system with the traditional parameters. The results are shown in Fig. 1. Figure 1(a) shows the three largest Lyapunov exponents of the autonomous RC plotted as a function of $\lambda_{max}^{(r)}$. The Lyapunov exponents of the true Lorenz system are shown as horizontal gray lines. The agreement between the first two Lyapunov exponents of the autonomous RC Lyapunov exponents and the target system Lyapunov exponents is excellent in all cases. For the third Lyapunov exponent (purple triangles), the agreement is excellent when $\lambda_{max}^{(r)}$ is very negative. However, as $\lambda_{max}^{(r)}$ increases and approaches the negative LE of the target system, the disagreement increases significantly. The reason for this disagreement is described in Section \ref{sec:apptoRC}: Once $\lambda_{max}^{(r)}$ has increased such that it is greater than the negative LE of the target system, the reservoir dynamics are now too slow to capture the negative LE of the target system, and the $\lambda_{max}^{(r)}$ replaces the negative LE of the target system as the third-largest LE of the autonomous RC. Evidence for this is provided by the fact that the $\lambda^{(RC)}_3$ mostly lie along the curve $\lambda = \max(\lambda_3^{(Lorenz)},\lambda_{max}^{(r)})$, as predicted in Section \ref{sec:apptoRC}.

Figure 1(b) shows that $\lambda_{max}^{(r)}$ is strongly correlated with spectral radius. Variations for a fixed spectral radius are due to the different input strengths and reservoir network topologies. For confirmation, Fig. 1(c) shows $\lambda_{i}^{(RC)}$ for $i\in\{1,2,3\}$ as a function of the reservoir spectral radius; one sees that the Lyapunov exponent estimation becomes unreliable for spectral radii greater than about 0.8. The large spread in $\lambda^{(RC)}_3$ for larger values of spectral radius is due to the fact that, at these values of spectral radius, $\lambda^{(RC)}_3$ is determined by $\lambda_{max}^{(r)}$, which is not fully determined by (only strongly correlated with) the spectral radius.

We believe that this effect is the reason that previous efforts \cite{pathak2017using} at attractor reconstruction using a large reservoir spectral radius have been unsuccessful at replicating the negative Lyapunov exponent of the Lorenz system.

We now consider the attractor reconstruction task on the Lorenz system with the modified parameters. The results are shown in Fig. 2, and are very similar to the case of the traditional Lorenz parameters. Figure 2(a) shows the three largest Lyapunov exponents of the autonomous reservoir computer $\lambda_i^{(RC)}$ plotted as a function of the maximal CLE of the driven reservoir $\lambda_{max}^{(r)}$. The Lyapunov exponents of the true Lorenz system are shown as horizontal gray lines. The agreement between the first two Lyapunov exponents of the autonomous RC and the first two target system Lyapunov exponents is excellent in all cases. For the third Lyapunov exponent (purple triangles), the agreement is excellent when $\lambda_{max}^{(r)}$ is very negative. However, as $\lambda_{max}^{(r)}$ increases, the disagreement increases significantly. The reason for this disagreement is described in Section \ref{sec:apptoRC}: Once $\lambda_{max}^{(r)}$ has increased such that it is greater than the negative LE of the target system, the reservoir dynamics are too slow to capture the negative LE of the target system, and $\lambda_{max}^{(r)}$ replaces the negative LE of the target system as the $\lambda_{3}^{(RC)}$. Evidence for this is provided by the fact that the $\lambda_3^{(RC)}$ mostly lie along the curve $\lambda = \max(\lambda_3^{(Lorenz)},\lambda_{max}^{(r)})$, as predicted in Section \ref{sec:apptoRC}.

There appear to be no deleterious effects of having a very negative maximal CLE, corresponding to a short reservoir memory, for this task. Figure 2(b) shows that the maximal CLE is strongly correlated with spectral radius. Variations for a fixed spectral radius are due to the different input strengths and reservoir network topologies.

Figure 2(c) shows $\lambda_i^{(RC)}$ as a function of the reservoir spectral radius; one sees that the Lyapunov exponent estimation becomes unreliable for spectral radii greater than about 0.7. The large spread in $\lambda^{(RC)}_3$ for larger values of spectral radius is due to the fact that, at these values of spectral radius, $\lambda^{(RC)}_3$ is determined by $\lambda_{max}^{(r)}$, which is not fully determined by (only strongly correlated with) the spectral radius.

\begin{figure}
    \centering
    \begin{subfigure}[b]{0.45\textwidth}
    \includegraphics[width=\textwidth]{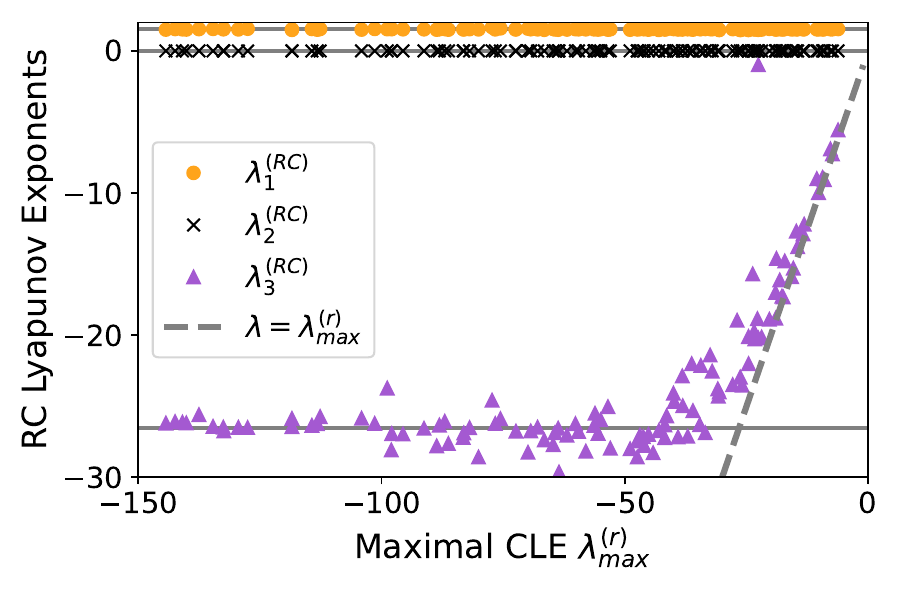}
    \caption{}
    \end{subfigure}
    \begin{subfigure}[b]{0.45\textwidth}
    \includegraphics[width=\textwidth]{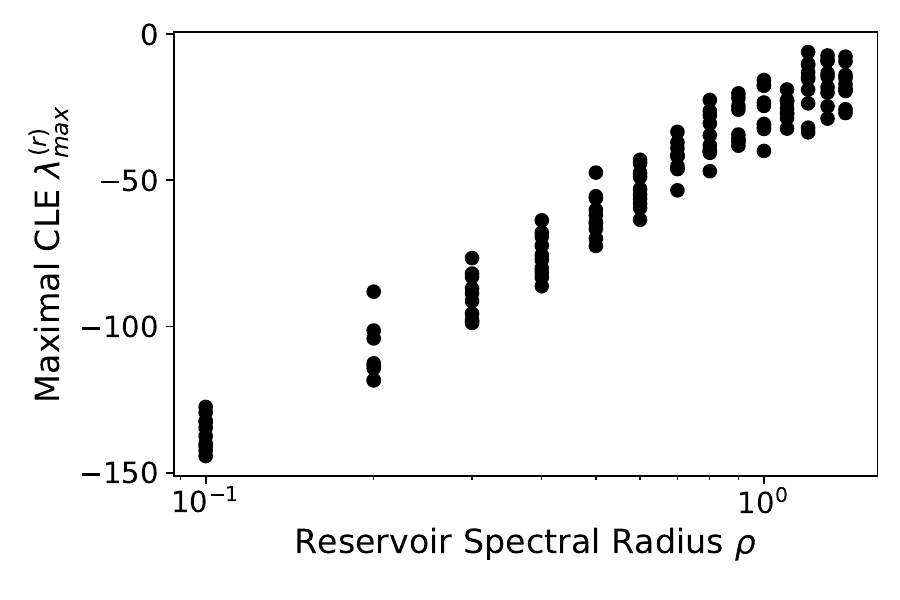}
    \caption{}
    \end{subfigure}
    \begin{subfigure}[b]{0.45\textwidth}
    \includegraphics[width=\textwidth]{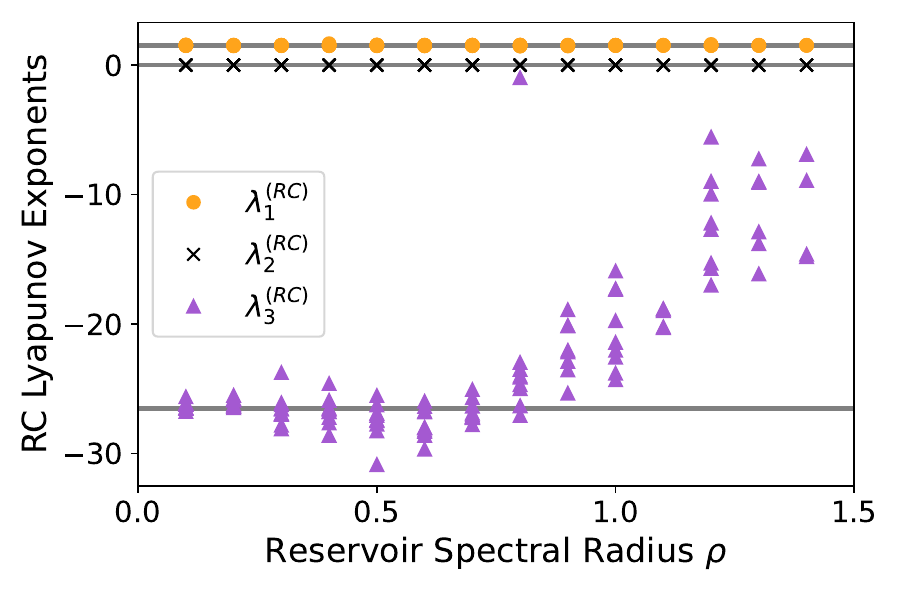}
    \caption{}
    \end{subfigure}
    \caption{Lorenz system with $\sigma=20$, $\beta=4$, and $\rho=45.92$. (a) Three largest Lyapunov exponents of the trained autonomous RC $\lambda_i^{(RC)}$ vs the maximal CLE of the driven reservoir $\lambda_{max}^{(r)}$. The Lyapunov exponents of the target Lorenz system are indicated as solid gray lines. The line $\lambda=\lambda_{max}^{(r)}$ is shown as a dashed gray line. (b) Maximal CLE of the driven reservoir $\lambda_{max}^{(r)}$ vs the reservoir spectral radius. (c) Three largest Lyapunov exponents of the trained autonomous RC $\lambda^{(RC)}$ vs the reservoir spectral radius. }
\end{figure}

We now consider the effect of $\lambda_{max}^{(r)}$ on the Kaplan-Yorke dimension of the autonomous RC, shown in Fig. 3. As $\lambda_{max}^{(r)}$ nears and surpasses the negative Lorenz Lyapunov exponent, the Kaplan-Yorke dimension increases, as expected (Section \ref{sec:filtering}). Therefore, we find that a large $\lambda_{max}^{(r)}$ (or large reservoir spectral radius) results in poor attractor reconstruction, as quantified by replication of the entire target Lyapunov spectrum.

\begin{figure}
    \centering
    \begin{subfigure}[b]{0.45\textwidth}
    \includegraphics[width=\textwidth]{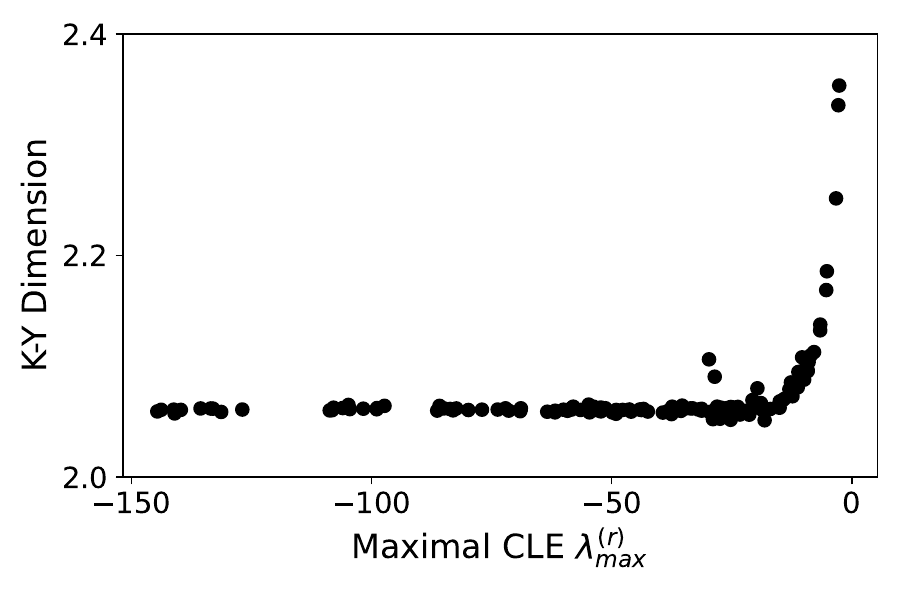}
    \caption{}
    \end{subfigure}
    \begin{subfigure}[b]{0.45\textwidth}
    \includegraphics[width=\textwidth]{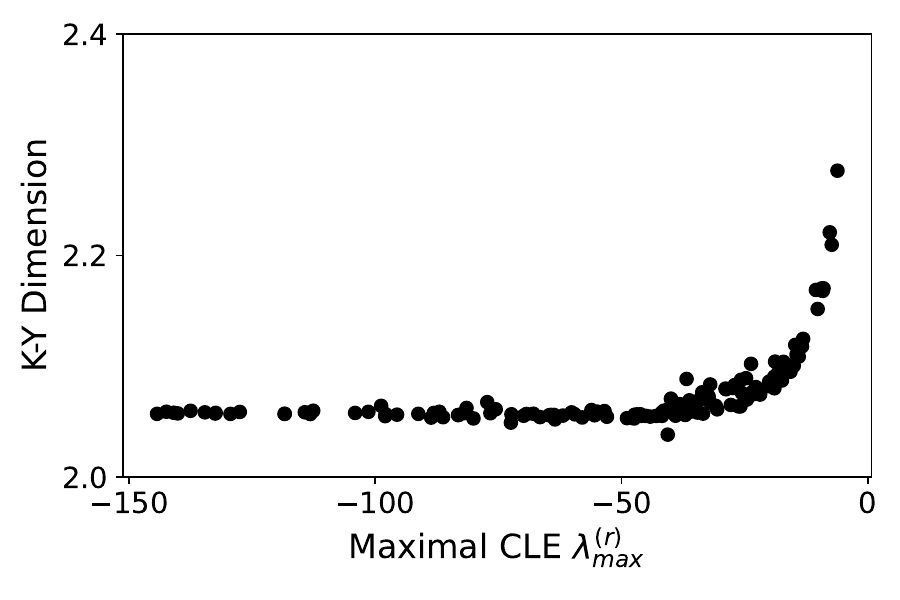}
    \caption{}
    \end{subfigure}
    \caption{Kaplan-Yorke dimension of the trained autonomous RC vs maximal CLE of the driven reservoir before training for the Lorenz system with (a) $\sigma=10$, $\beta=8/3$ and $\rho=28$ and (b) $\sigma=20$, $\beta=4$, and $\rho=45.92$. In both cases, the Kaplan-Yorke dimension of the RC increases sharply as the maximal CLE of the reservoir $\lambda_{max}^{(r)}$ approaches zero (i.e., as $\rho$ increases), indicating that the autonomous RC does not accurately replicate the Lorenz attractor geometry when $\rho$ is near unity. }
\end{figure}

\section{Results: Qi System}
\subsection{The Qi Equations}
The Qi system \cite{qi2005four} is a well-characterized four-dimensional chaotic system with a butterfly attractor. The Qi system is described by
\begin{equation}
    \label{eq:Qi}
    \begin{array}{l}
    \frac{dx_1}{dt} = p_1(x_2-x_1) + x_2x_3x_4 +a\xi_1(t)\\
    \frac{dx_2}{dt} = p_2(x_1+x_2)+ x_1x_3x_4 + a\xi_2(t) \\
    \frac{dx_3}{dt} = -p_3x_3 + x_1x_2x_4+a\xi_3(t) \\
    \frac{dx_4}{dt} = -p_4x_4 + x_1x_2x_3+a\xi_4(t).
\end{array}
\end{equation}
As with the Lorenz system, we choose a noise strength $a=0.01$. In this work, we choose $p_1=35$, $p_2=10$, $p_3=1$ and $p_4=10$ since these parameter values are known to result in a chaotic system with one positive Lyapunov exponent, and two well-separated negative Lyapunov exponents (along with a zero exponent) \cite{qi2005four}. For these parameters, we compute a Lyapunov spectrum of 3.26, 0.00, -4.14, and -35.12. These well-separated negative Lyapunov exponents create a more strenuous test of our hypothesis that $\lambda_{max}^{(r)}$ (which is strongly dependent on the reservoir spectral radius) places a limitation on the minimum Lyapunov exponent that the autonomous RC can reproduce.

The dynamical noise $\xi(t)$ is Gaussian white noise with unit standard deviation and the different $\xi_i$ are independent. We choose the dynamical noise strength $a=0.01$; dynamical noise is known to help with the RC training in some cases \cite{banerjee2019using, rohm2021model,banerjee2021machine}. To generate the training data, we integrate Eq. \ref{eq:Qi} using Honeycutt's second order Runge-Kutta method \cite{honeycutt1992stochastic} with a time step of 0.0001. To obtain the RC input data, we downsample the time series so that $\tau=0.01$. All four variables from the Qi system are used as inputs to the reservoir.

\subsection{Attractor reconstruction}

\begin{figure}[h!]
    \centering
    \begin{subfigure}[b]{0.4\textwidth}
    \includegraphics[width=\textwidth]{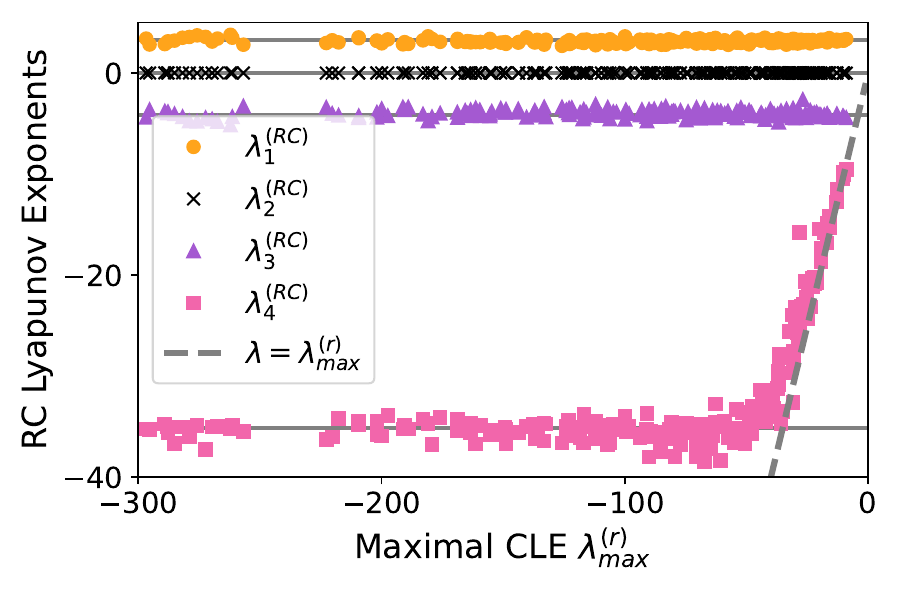}
    \caption{}
    \end{subfigure}
    \begin{subfigure}[b]{0.4\textwidth}
    \includegraphics[width=\textwidth]{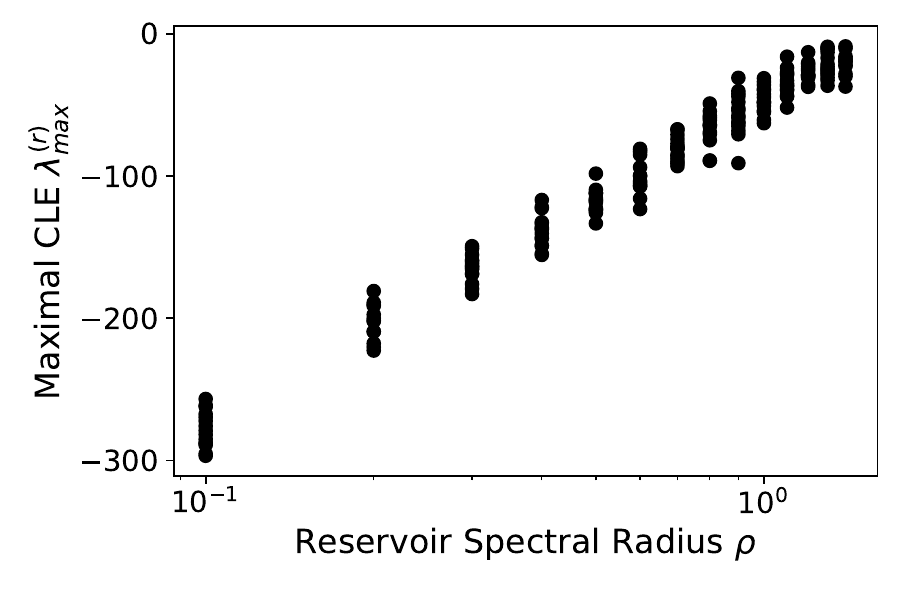}
    \caption{}
    \end{subfigure}
     \begin{subfigure}[b]{0.4\textwidth}
    \includegraphics[width=\textwidth]{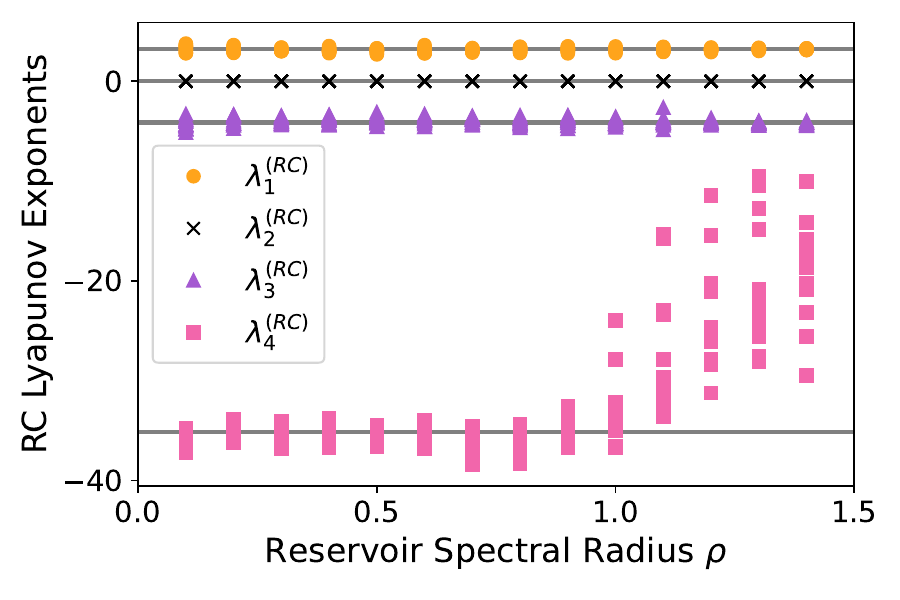}
    \caption{}
    \end{subfigure}
    \caption{Qi system with $p_1=35$, $p_2=10$, $p_3=1$ and $p_4=10$. (a) Four largest Lyapunov exponents of the trained autonomous RC $\lambda_i^{(RC)}$ vs the maximal CLE of the driven reservoir $\lambda_{max}^{(r)}$. The Lyapunov exponents of the target Qi system are indicated as solid gray lines. The line $\lambda=\lambda_{max}^{(r)}$ is shown as a dashed gray line. (b) Maximal CLE of the driven reservoir $\lambda_{max}^{(r)}$ vs the reservoir spectral radius. (c) Four largest Lyapunov exponents of the trained autonomous RC $\lambda_i^{(RC)}$ vs the reservoir spectral radius. }
\end{figure}

For each value of the reservoir spectral radius, we create, train, and test 1000 RCs with $N=400$ nodes and with $\mathbf{P}(\mathbf{x})=\mathbf{x}$. We use 400 nodes. % because the dimension of the Qi system is 4 (rather than 3 for the Lorenz system). 
The RCs differ in their input strengths $\sigma_{in}$ (chosen randomly in the range (0,1]), network topologies (Erd{\"o}s-Renyi random networks with average degree 6), input matrices, and bias vectors. 
We use 40000 training time steps and a ridge regression parameter of $10^{-8}$ is used. Again, the ten ``best'' attractor reconstructions are retained, where the attractor reconstructions are ranked according to the mean square error between the power spectra of the reconstructed $x_1$ variable and of the true $x_1$ variable over 20000 time steps. The Lyapunov spectra of the trained, autonomous RCs $\lambda_{i}^{(RC)}$ are computed using 100000 time steps. Additionally, the CLE spectra of the driven reservoirs $\lambda_{i}^{(r)}$ are computed over the 40000 training time steps for these retained reservoirs.

The results of the RC trained on the Qi system are shown in Fig. 4. Figure 4(a) shows the $\lambda_{i}^{(RC)}$ for $i\in\{1,2,3,4\}$ plotted as a function of the $\lambda_{max}^{(r)}$. The Lyapunov exponents of the target system are shown as horizontal gray lines. The agreement between the first three Lyapunov exponents of the autonomous RC and the target system Lyapunov exponents is excellent in all cases. For the fourth Lyapunov exponent (pink squares), the agreement is excellent when $\lambda_{max}^{(r)}$ is very negative. However, as $\lambda_{max}^{(r)}$ increases, the disagreement increases significantly. The reason for this disagreement is described in Section \ref{sec:apptoRC}: Once $\lambda_{max}^{(r)}$ has increased such that it is greater than the most negative LE of the target system, the reservoir dynamics are now too slow to capture the negative LE of the target system, and $\lambda_{max}^{(r)}$ replaces the negative LE of the target system as $\lambda_{4}^{(RC)}$. Evidence for this is provided by the fact that the $\lambda_{4}^{(RC)}$ mostly lie along the curve $\lambda = \max(\lambda_4^{(Qi)},\lambda_{max}^{(r)})$, as predicted in Section \ref{sec:apptoRC}. Again, there appear to be no deleterious effects of having a very negative $\lambda_{max}^{(r)}$, corresponding to a short reservoir memory, for this task. 

Figure 4(b) shows that $\lambda_{max}^{(r)}$ is strongly correlated with spectral radius. Variations for a fixed spectral radius are due to the different input strengths and reservoir network topologies. For confirmation of this, Fig. 4(c) shows $\lambda_{i}^{(RC)}$ for $i\in\{1,2,3,4\}$ as a function of the reservoir spectral radius; one sees that the Lyapunov exponent estimation becomes unreliable for spectral radii greater than about 0.9. The large spread in $\lambda_4^{(RC)}$ for larger values of spectral radius is due to the fact that, at these values of spectral radius, $\lambda_4^{(RC)}$ is determined by $\lambda_{max}^{(r)}$, which is not fully determined by (only strongly correlated with) the spectral radius.

We now consider the effect of $\lambda_{max}^{(r)}$ (or the reservoir spectral radius) on the Kaplan-Yorke dimension, shown in Fig. 5. As $\lambda_{max}^{(r)}$ nears and surpasses the most negative Lyapunov exponent of the Qi system, the reservoir Kaplan-Yorke dimension does not change. This is expected, because the Kaplan-Yorke dimension of this target system depends on only the first three Lyapunov exponents. Therefore, in this case a large $\lambda_{max}^{(r)}$ (large spectral radius) results in poor estimation of the most negative Lyapunov exponent, but does not impact the dimensionality of the reconstructed attractor.

\begin{figure}
    \centering
    \includegraphics[width=0.5\textwidth]{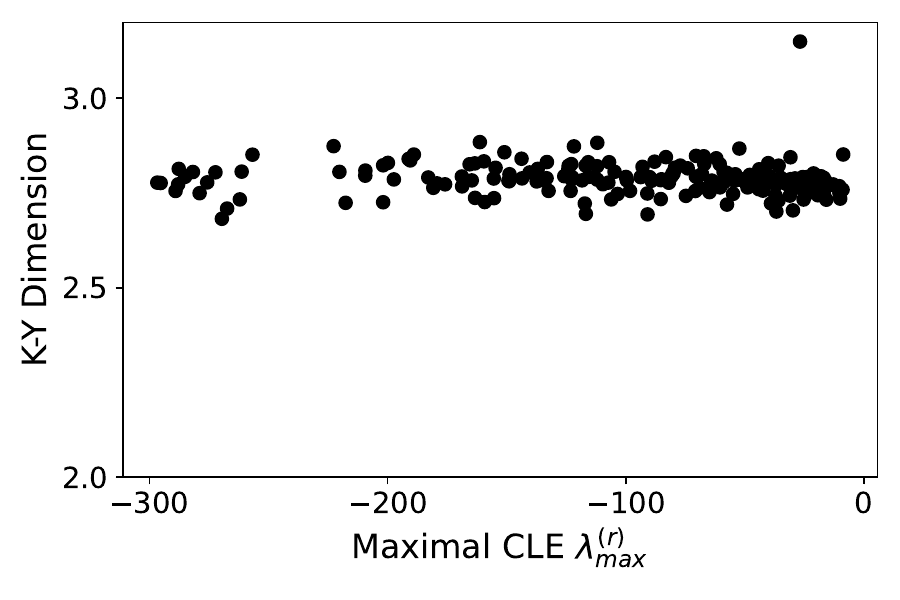}
    \caption{Kaplan-Yorke dimension of the trained autonomous RC vs largest CLE $\lambda_{max}^{(r)}$ of the driven reservoir for the Qi system. For the Qi system, the Kaplan-Yorke dimension of the RC does not depend much on the largest CLE because it is determined solely by the three largest Lyapunov exponents, which are all relatively small in absolute value.}
\end{figure}

\section{Discussion}

A comparison of the results on the Lorenz system and the Qi system show similar behavior: In the attractor reconstruction task, a reservoir computer is able to reproduce only the Lyapunov exponents of the target system that are smaller in magnitude than than maximal conditional Lyapunov exponent of the untrained reservoir $\lambda_{max}^{(r)}$. Further, $\lambda_{max}^{(r)}$ is strongly correlated with the spectral radius of the reservoir adjacency matrix, which provides an effective way to tune $\lambda_{max}^{(r)}$. However, it is important to note that $\lambda_{max}^{(r)}$ also depends on the target system (i.e., the drive system); this can be seen by comparing Figs. 2b and 4b. For a given spectral radius, the $\lambda_{max}^{(r)}$ for the reservoir driven by the Qi system is much more negative than $\lambda_{max}^{(r)}$ for the reservoir driven by the Lorenz system. This also manifests itself in the ``threshold'' spectral radius--the largest spectral radius for which $\lambda^{(RC)}$ provides a reliable estimate of $\lambda_{min}^{(drive)}$ (Figs. 2c and 4c): Even though the Qi system has a more negative Lyapunov exponent (-35.12) than the Lorenz system (-26.5), the threshold spectral radius of for the Qi system (0.9) is significantly larger than that of the Lorenz system (0.7). This confirms that the maximal CLE of the reservoir $\lambda_{max}^{(r)}$ is a more reliable predictor of RC performance than the spectral radius.

Our findings suggest that reservoirs with a small spectral radius are better for the attractor reconstruction task, as quantified by Lyapunov exponent and dimension estimation. Recent findings also suggest that reservoirs with small spectral radius perform better than those with large spectral radius in terms of the short-term accuracy and long-term stability of predictions \cite{jaurigue2024chaotic}; whether these results are related is a subject of ongoing research. Practically, having a small $\lambda_{max}^{(r)}$ has the additional benefit of a shorter warm up time than a reservoir with a longer memory \cite{lu2018attractor}.

These benefits of reservoirs with small CLEs provoke the question: Can one just set the reservoir spectral radius to zero, such that the reservoir has no memory and $\lambda_{max}^{(r)}=-\infty$? This special case of a reservoir computer is often called an extreme learning machine (ELM) \cite{huang2006extreme}. We have found that one can obtain good performance at the attractor reconstruction and Lyapunov exponent estimation tasks using an ELM in which $W^{in}$ is an $N\times\mathcal{D}$ random matrix. However, recent work \cite{jaurigue2024chaotic} finds that a small but non-zero spectral radius seems to be optimal for the long-term stability of the prediction of an autonomous RC, at least in the case of the Lorenz system.

%***********The dimension increase does not occur for finite impulse response (FIR) filters \cite{sauer1997dimensions}. ELM/NVAR are nonlinear FIR filters...
%*******In discussion/conclusion, make a note about how NVAR \cite{gauthier2021next} is not IIR, and so may not have these issues? Also ELM is not IIR. But NVAR has other issues, including stability if you get the nonlinearity wrong.

An important question for future work is whether the relationship between the maximal CLE of the driven reservoir and the Lyapunov exponents of the autonomous RC hold in the case where the RC is trained on only a single input variable (for example, the Lorenz $x$ variable only) or on a system with a time delay (for example, the Mackey-Glass system \cite{mackey1977oscillation}). In this case, the reservoir must create an embedding \cite{duan2023embedding}, and so the reservoir may require some memory to complete this task. As a result, there may be a tradeoff between having enough memory (which requires a large enough maximal CLE) for the embedding and having a small enough maximal CLE for the autonomous RC to reproduce the target negative Lyapunov exponents. It is possible that adding time shifts on the reservoir output \cite{del2021reservoir,carroll2022time,duan2023embedding} or input \cite{kobayashi2021dynamical} can increase the memory without modifying the maximal CLE of the reservoir; however, the addition of time shifts will introduce time-delayed feedback loops into the autonomous reservoir computer, which can complicate the Lyapunov spectrum computation and even destabilize the dynamics.

\section{Conclusions}

We demonstrated that RCs perform best at the attractor reconstruction task when the maximal CLE of the reservoir to the training signal is significantly more negative than the most negative Lyapunov exponent of the target system (or, at least, the most negative Lyapunov exponent that one wishes to reproduce). Attractor reconstruction is defined here as replicating the ergodic properties--here exemplified by the Lyapunov spectra and information dimension--of the target dynamical system \cite{platt2021robust,platt2022systematic}. We argued this is related to the properties of the generalized synchronization function \cite{lu2018attractor, davies1997reconstructing} and to the idea that a filter driven by a chaotic system can show an increased fractal dimension if the filter's maximal CLE is larger than the drive system's negative Lyapunov exponents \cite{badii1988dimension}. We also found that the maximal CLE of the reservoir is strongly correlated with the spectral radius of the reservoir. 

Our results show how the generalized synchronization behavior of the driven reservoir plays a key role in determining the dynamical behavior of the autonomous RC. While the link between generalized synchronization and reservoir computing has long been understood, we show a quantitative relationship between the rate at which an untrained reservoir approaches generalized synchronization and the performance of the trained RC at the attractor reconstruction task.

Practically, our results can assist the practitioner of reservoir computing. First, our findings suggest that often a RC with a spectral radius significantly smaller than unity performs best at the attractor reconstruction task. Second, when attempting to use reservoir computing to estimate the Lyapunov exponents of an unknown dynamical system, one can compute not only the Lyapunov exponents of the autonomous RC, but also the maximal CLE of the reservoir driven by the training signal. One should trust only the negative Lyapunov exponents from the autonomous RC that are significantly larger than the maximal CLE of the driven reservoir. This is a relatively inexpensive check, since only the maximal CLE must be computed.

%A more rigorous test for attractor reconstruction reliability is suggested by Figs. 2a and 4a. One can attempt attractor reconstruction for RCs with many different maximal CLEs (which can practically be tuned by the spectral radius). If the Lyapunov spectrum remains approximately constant The drawback is that this can be quite computationally expensive.

\textbf{Acknowledgment} JDH wishes to acknowledge Lou Pecora, Tom Carroll, and Andrew Flynn for insightful discussions. This work was supported by the US Naval Research Laboratory's Base Program.

%\nocite{*}
%\bibliographystyle{unsrt}
\bibliography{sample}

\end{document}